\documentclass[letterpaper, 10 pt, conference]{ieeeconf}  

\IEEEoverridecommandlockouts                              

\usepackage{amsmath} 
\usepackage{amssymb}  
\usepackage{adjustbox}

\usepackage{colortbl}       
\usepackage{pifont}         
\usepackage{multirow}       
\usepackage{booktabs}       

\usepackage[table]{xcolor}
\definecolor{waymolgray}{HTML}{F0F0F0}  
\definecolor{waymollblue}{HTML}{CCE4FF} 
\definecolor{waymoblue}{HTML}{0077FF}
\definecolor{waymolblue}{HTML}{99B7FF}  

\title{\LARGE \bf
LEGO-Motion\raisebox{-0.1\height}{\includegraphics[width=0.04\linewidth]{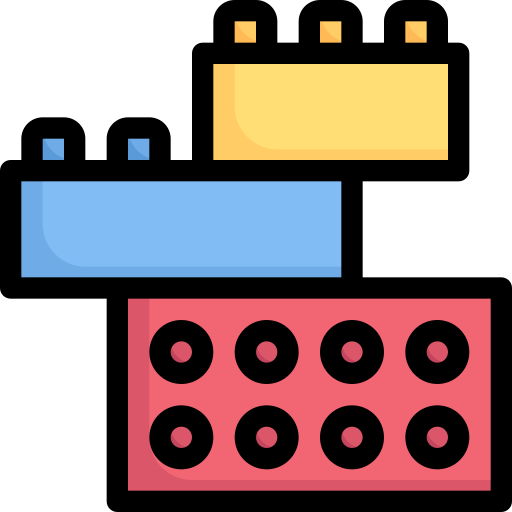}}: Learning-Enhanced Grids with Occupancy Instance Modeling for Class-Agnostic Motion Prediction}

\author{Kangan Qian$^{1,2}$, Jinyu Miao$^{1}$, Ziang Luo$^{1}$, Zheng Fu$^{1}$, Jinchen Li$^{1}$, Yining Shi$^{1}$, \\Yunlong Wang$^{3}$, Kun Jiang$^{1,*}$, Mengmeng Yang$^{1}$, Diange Yang$^{1,*}$ %
\thanks{*This work was supported in part by National Key Research and Development Program of China(2024YFB2505803,2024YFB2505802)}
\thanks{$^{1}$Kangan Qian, Jinyu Miao, Ziang Luo, Zheng Fu, Jinchen Li, Yining Shi, Kun Jiang, Mengmeng Yang, and Diange Yang are with the School of Vehicle and Mobility, Tsinghua University, Beijing, China.}%
\thanks{Kangan Qian was with AI2Robotics during his internship in Beijing, China.}
\thanks{$^{3}$Yunlong Wang is with AD Division of NIO Inc.,
China.}%
\thanks{$^{*}$Corresponding author: Diange Yang and Kun Jiang.}
}

\begin{document}
\maketitle
\thispagestyle{empty}
\pagestyle{empty}

\begin{abstract}

Accurate and reliable spatial and motion information plays a pivotal role in autonomous driving systems. However, object-level perception models struggle with handling open scenario categories and lack precise intrinsic geometry. On the other hand, occupancy-based class-agnostic methods excel in representing scenes but fail to ensure physics consistency and ignore the importance of interactions between traffic participants, hindering the model's ability to learn accurate and reliable motion. In this paper, we introduce a novel occupancy-instance modeling framework for class-agnostic motion prediction tasks, named LEGO-Motion, which incorporates instance features into Bird's Eye View (BEV) space. Our model comprises (1) a BEV encoder, (2) an Interaction-Augmented Instance Encoder, and (3) an Instance-Enhanced BEV Encoder, improving both interaction relationships and physics consistency within the model, thereby ensuring a more accurate and robust understanding of the environment. Extensive experiments on the nuScenes dataset demonstrate that our method achieves state-of-the-art performance, outperforming existing approaches. Furthermore, the effectiveness of our framework is validated on the advanced FMCW LiDAR benchmark, showcasing its practical applicability and generalization capabilities. The code will be made publicly available to facilitate further research.


\end{abstract}

\section{Introduction}


Autonomous driving systems depend on an accurate perception of the surroundings, including spatial position information \cite{spatial}, semantic classification \cite{maskformer}, and motion behaviour  \cite{motion}, all of which are crucial for downstream planning tasks. The common paradigm for perception is object-centric methods \cite{pointrcnn, centerpoint, pointpillars}, which represent traffic participants as bounding boxes (bbox) and formulate their motion behaviour as trajectory prediction tasks. Despite their apparent advantages, these methods are not equipped to handle unexpected categories, which are critical for system safety and which have not been seen in the training set due to the reliance on detection-tracking-prediction pipelines. This poses a threat to intelligent vehicles in open 3D scenarios. Moreover, the limitations of BBox are becoming increasingly evident as the demand for enhanced perception accuracy continues to increase. BBox-based representations are unable to capture the precise details of an object's shape, particularly for objects with irregular geometries\cite{ObjectAugments}. Occupancy-based representation has emerged as a promising alternative for perception systems \cite{streamingflow, LSTM-ED}. It discretises the scene into unified and regular grids, which flexibly represent different kinds of objects with diverse geometries. However, detecting and predicting the class and motion of 3D occupancy grids requires a high computation, which does not meet the real-time prediction needs of self-driving systems \cite{grid-centric}. 

\begin{figure}[!t]
\centering
\includegraphics[scale=0.36]{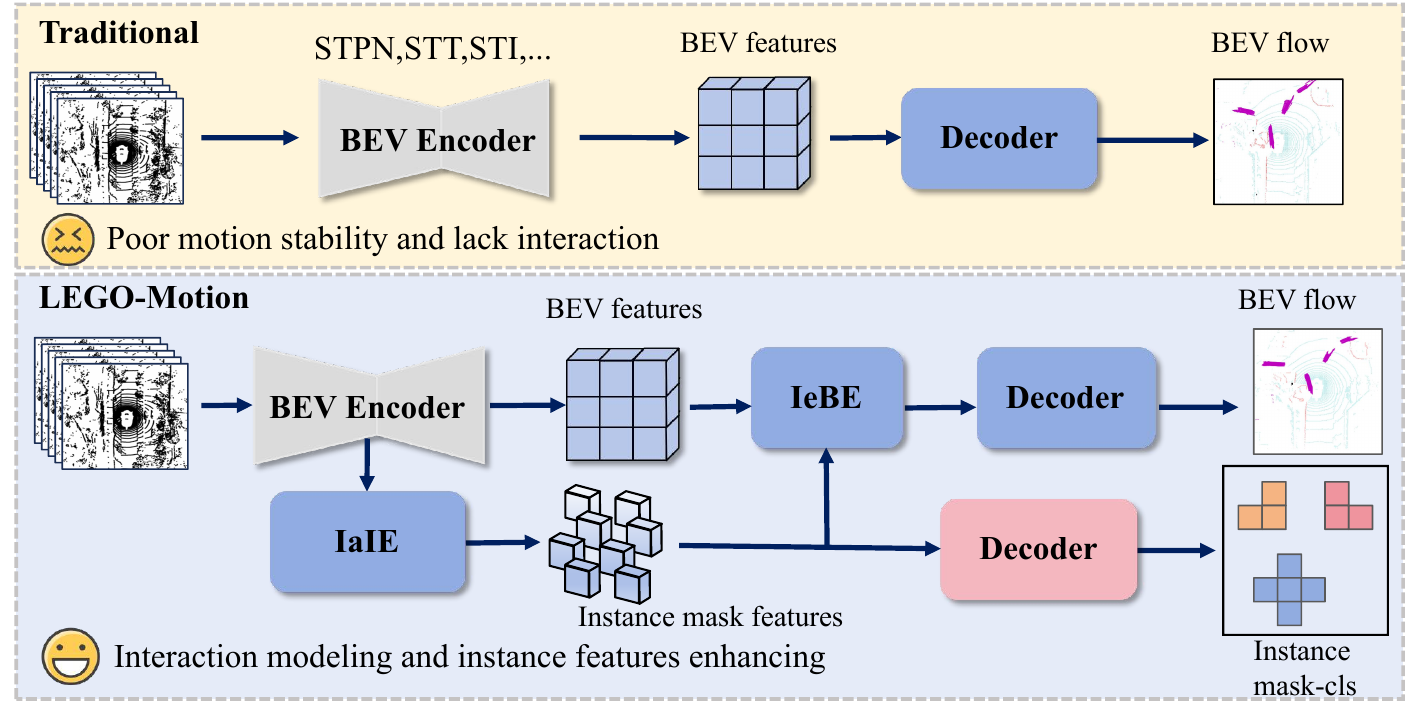}
\caption{Comparison between LEGO-Motion and previous motion prediction framework. \textbf{Top row:} Previous framework, which adapts encoder-decoder pipeline to capture and predict motion fields. \textbf{Bottom:} Our proposed LEGO-Motion.}
\label{figurelabel}
\end{figure}
   
Recently, class-agnostic with BEV occupancy grid representation methods have provided an attractive solution by estimating the semantic class and dense motion field through a Bird’s Eye View (BEV) occupancy grid approach. MotionNet \cite{motionnet} is the pioneering work to process 3D occupancy grids and jointly perform perception and motion prediction on 2D BEV maps, followed by BE-STI \cite{BE-STI} and ST-Transformer \cite{ST-Transformer} showing significant improvements with enhanced backbones. Nonetheless, despite the encouraging nature of grid-based prediction, physical coherence is not guaranteed. Specifically, the presence of inconsistent flows within a single object can be observed in the results of visualisations. The per-pixel design fails to consider instance consistency, which is a source of confusion for model learning. A second problem is the absence of interaction between traffic participants, which is important for the prediction of motion behaviour. The grid-based representation makes it technically difficult to model the interaction between grids. 
As posited by some researchers \cite{motionnet, self-supervised}, the proposal of a spatial consistency loss function is intended to ensure the same motion prediction within a single instance. However, the improvements are negligible, and the process of setting the hyper-parameters is time-consuming.

Our key observation stems from the complementary nature of two distinct paradigms: detection-tracking-prediction frameworks (object-centric representations) and class-agnostic occupancy methods (grid-based representations). This dichotomy mirrors human cognitive capabilities in the simultaneous processing of holistic object structures and granular spatial details. This is analogous to the manner in which LEGO systems seamlessly integrate predefined building blocks with customisable granular elements. Drawing inspiration from the mask classification paradigm in semantic segmentation \cite{SP, SDS}, where binary masks associate with global class labels, a natural question emerges: \textit{Can we leverage instance-level features to enhance grid-level representations, thereby improving the performance of class-agnostic tasks?}


\begin{table}[ht]
    \centering
    \caption{Motion prediction mean error (m) comparison on nuScenes.}
    \begin{tabular}{l c c c}
        \toprule
        \textbf{Method} & \textbf{Static$\downarrow$} & \textbf{Speed $\leq$ 5 m/s$\downarrow$} & \textbf{Speed $>$ 5 m/s$\downarrow$} \\
        \midrule
        Base [35] & 0.0239 & 0.2467 & 1.0109 \\
        Base + GT$_{\text{cls-mask}}$ & 0.0184 & 0.2117 & 0.8369 \\
        \bottomrule
    \end{tabular}
    
    \label{tab:toy_example}
\end{table}
To address this question, we first examine a toy example. As shown in Tab.~\ref{tab:toy_example}, we provide MotionNet \cite{motionnet} with semantic-included instance mask ground truth (GT) as additional input. The process is initiated with the conversion of the pixel instance GT into mask representations, with the segmentation category serving as an additional vector input. This approach outperforms the baseline to some extent, confirming that high-quality semantic-included instance information positively impacts motion prediction tasks.


In this paper, a simple yet efficient \textbf{LEGO-Motion} framework is proposed, incorporating instance-level feature learning to enhance grids for occupancy-based motion prediction. In light of the absence of interaction with grid-based representation, an Interaction-Augmented Instance Encoder (IaIE) with an attention mechanism is proposed to extract social interaction relationships. Another concern regarding the proposed framework pertains to the full injection of the instance information generated by the instance encoder. To this end, we propose an Instance-Augmented BEV Encoder (IaBE), which captures instance priors fused with the BEV feature map, thereby facilitating robust class-agnostic motion prediction tasks in 3D environments. The contributions of our framework can be summarised as follows:

\begin{itemize}
    \item We propose \textbf{LEGO-Motion}, a novel framework that integrates instance-level feature learning into occupancy-based motion prediction. By leveraging detailed object interactions and instance priors, LEGO-Motion enhances traditional pipelines, providing more accurate and consistent predictions.
    
    \item We introduce the \textbf{Interaction-augmented Instance Encoder (IaIE)} and the \textbf{Instance-enhanced BEV Encoder (IeBE)}. These components enhance the understanding of object interactions and seamlessly integrate instance priors into the BEV feature maps.

    \item Extensive experiments on the \textbf{nuScenes} dataset \cite{nuScenes} and our private FMCW LiDAR benchmark demonstrate that \textbf{LEGO-Motion} enhances previous state-of-the-art methods. Our approach shows significant improvements in motion prediction accuracy and robustness across various challenging scenarios.
\end{itemize}

\section{Related Work}
\subsection{Motion Prediction in BEV grid cells}
Motion prediction aims to extract traffic participants' future trajectories from past observations \cite{Efficient}. Recently, BEV grid-based methods have gained popularity due to their flexibility in representing the zigzag shapes of objects and their independence from upstream detection and tracking results\cite{occupancy-flow}. These approaches are more robust in open scenarios, effectively perceiving objects not seen in the training set. MotionNet\cite{motionnet} is the pioneering work in this field, jointly performing spatial perception, cell classification and motion prediction tasks. PillarMotion\cite{pillarmotion} introduces a cross-sensor self-supervision framework to train MotionNet using optical flow supervision from RGB images. BE-STI\cite{BE-STI} enhances motion prediction performance by leveraging semantic tasks with the help of spatial and temporal bidirectional-enhanced encoders. ST-Transformer\cite{ST-Transformer} utilizes an attention mechanism, which is beneficial for feature extraction within voxelized point clouds. PriorMotion\cite{priormotion} proposes a novel framework to exploit priors from the motion field, effectively extracting motion patterns to improve prediction performance. 
Further work focuses on self-supervised methods to reduce human effort\cite{weak-supervised, self-supervised, semi-supervised, cross-modal}. In our work, we focus on instance-level feature extraction and its enhancement of BEV feature maps.

\subsection{Mask Prediction}
Mask prediction has gained widespread popularity in the field of computer vision, where it is used for instance-level segmentation tasks \cite{maskformer}. Mask R-CNN \cite{maskrcnn} introduces a global classifier to perform mask proposal classification for instance segmentation. DETR \cite{DETR} further incorporates the Transformer architecture \cite{transformer} for panoptic segmentation. While these methods support flexible output of various mask predictions, they require bounding box predictions, which may limit their applicability in semantic segmentation. MaskFormer \cite{maskformer} seamlessly converts any existing per-pixel classification model into a mask classification framework, thereby addressing both semantic and instance-level segmentation tasks in a unified manner. For BEV-based class-agnostic motion prediction tasks, integrating mask prediction with traditional pipelines remains an underexplored area.

\section{Methodology}
   \begin{figure*}[thpb]
      \centering
      \includegraphics[scale=0.5]{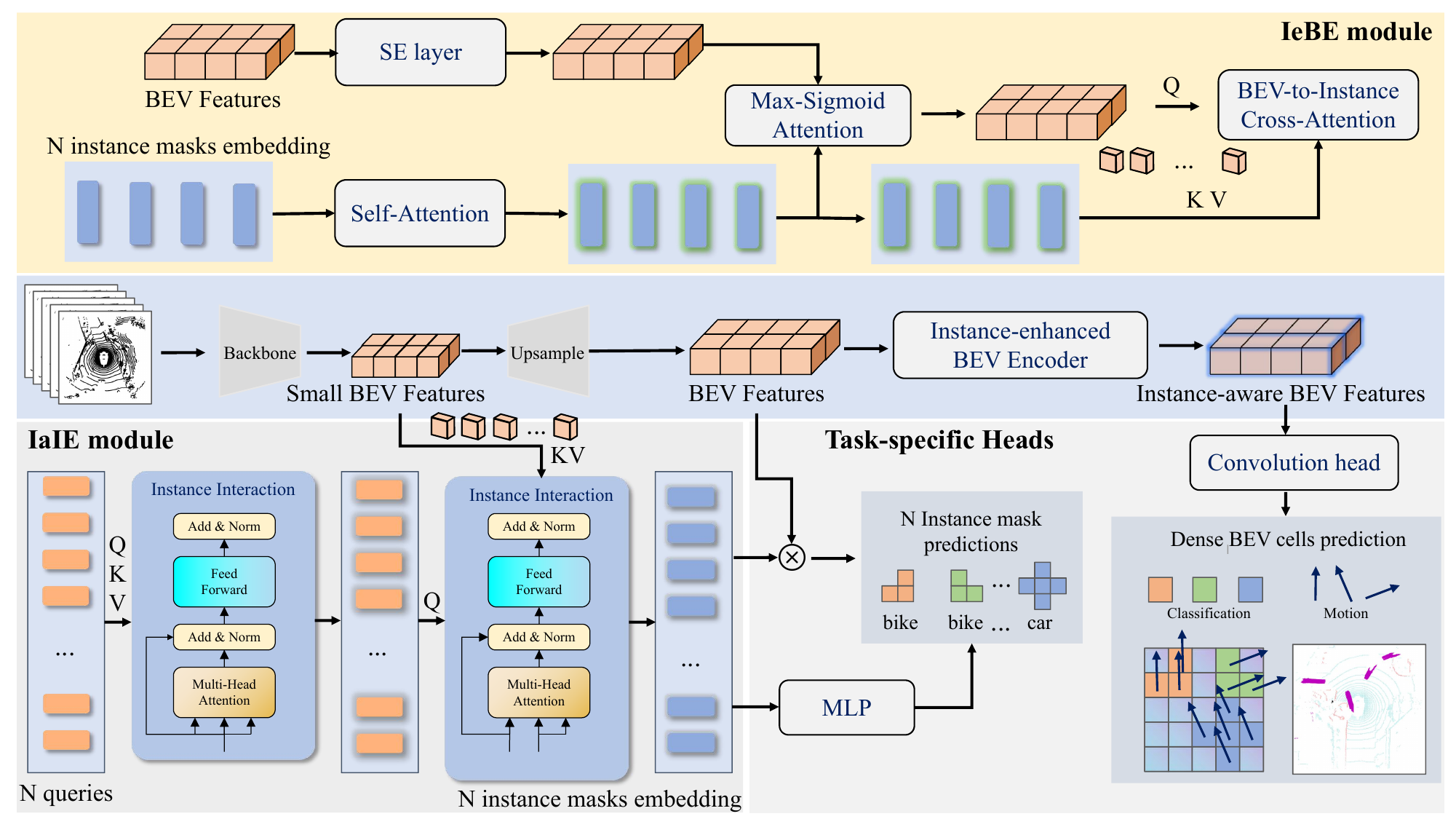}
      \caption{Architecture of learning-enhanced grids for with occupancy instance modeling(LEGO-Motion). \textbf{Top: }Instance-enhanced BEV Encoder. \textbf{Bottom left: }Interaction-augmented Instance Encoder. \textbf{Bottom right: }Task-specific Heads.}
      \label{fig:framework}
   \end{figure*}
\subsection{Problem Formulation}
Given a sequence of LiDAR point clouds $ P_t = \{\mathcal{P}_t^i\}_{i=1}^{N_t} $, we convert them into multi-frame occupancy-gridded point clouds $ \mathcal{V}_t \in \{0, 1\}^{H \times W \times C} $, where $ H $, $ W $, and $ C $ represent the voxel grid dimensions along the X, Y, and Z axes, respectively. Non-empty voxels are set to 1, while empty ones are set to 0. Our LEGO-Motion framework outputs per-pixel motion, classification, and state information, along with per-grid motion masks for instance modeling with semantic categories and motion states. The architecture consists of three key components:
(a) BEV Encoder;
(b) Interaction-Augmented Instance Encoder; and
(c) Instance-enhanced BEV Encoder.

\subsection{BEV Encoder} 
The BEV encoder takes multi-frame voxelized point clouds $ \mathcal{V}_t $ as input to generate a BEV feature map  $\mathcal{B}\in \mathbb{R}^{H \times W \times D}$, where $H \times W$ is the grid resolution and $D$ is the feature dimension. In particular, the BEV encoder in our framework can be flexibly chosen from various previous architectures, including the Spatial-Temporal Pyramid Network from MotionNet \cite{motionnet}, the transformer-based spatial and temporal attention network \cite{ST-Transformer}, spatially bidirectional enhanced encoder (TeSE and SeTE) from BE-STI \cite{BE-STI}.

\subsection{Interaction-augmented Instance Encoder}
Previous works \cite{motionnet, BE-STI, ST-Transformer} lack instance-level spatial coherence in grid-based motion prediction tasks, thereby overlooking potential future interactions between traffic participants. Inspired by the capabilities of self-attention layers and the Transformer architecture, our framework models the interaction relationships between different instances using an attention mechanism named the \textbf{Interaction-Augmented Instance Encoder}.

This module initializes $N$ instance queries $\mathcal{Q} \in \mathbb{R}^{N \times D}$, where $D$ is the dimension. These queries first interact through \textbf{self-attention} to model inter-instance relationships. The self-attention mechanism can be expressed as:


\begin{equation}
    \texttt{selfattn}(\mathcal{Q}, \mathcal{Q}) = \texttt{softmax}\left(\frac{\mathcal{Q} \mathcal{Q}^{\top}}{\sqrt{D}}\right) \mathcal{Q}
\end{equation}

Subsequently, we perform \textbf{cross-attention} between instance queries and the downsampled BEV features $\mathcal{B}^{'} \in \mathbb{R}^{H_c \times W_c \times D}$ to capture instance-to-BEV interactions:


\begin{equation}
    \texttt{crossattn}(\mathcal{Q}, \mathcal{B}') = \texttt{softmax}\left(\frac{\mathcal{Q}{(\mathcal{B}')}^{\top}}{\sqrt{D}}\right) \mathcal{B}'
\end{equation}

After these attention operations, the module generates $N$ instance mask embeddings $\mathcal{I} \in \mathbb{R}^{N \times D}$, which encapsulate both inter-instance relationships and instance-to-BEV interactions. These mask embeddings are subsequently utilized to enhance BEV feature maps and improve downstream tasks.

Specifically, the instance mask embeddings $\mathcal{I}$ are integrated into the upsampled BEV feature maps with original resolution $\mathcal{B} \in \mathbb{R}^{H \times W \times D}$, thereby enriching the spatial and semantic information available for downstream tasks. 
\begin{table*}[htbp]
\centering
\small
\caption{Comparison with State-of-the-Art Results on nuScenes. }
\vspace{-5pt}
\begin{tabular}{lccccccc}
\toprule
\multirow{2}{*}{Method} & \multirow{2}{*}{Backbone} & \multicolumn{2}{c}{Static} & \multicolumn{2}{c}{Speed $\leq$ 5m/s} & \multicolumn{2}{c}{Speed $>$ 5m/s} \\
\cmidrule(lr){3-4} \cmidrule(lr){5-6} \cmidrule(lr){7-8} & & Mean$\downarrow$ & Median$\downarrow$ & Mean$\downarrow$ & Median$\downarrow$ & Mean$\downarrow$ &  Median$\downarrow$ \\
\midrule
StaticModel & Rules & 0 & 0 & 0.6111 & 0.0971 & 8.6517 & 8.1412  \\
FlowNet3D\cite{flownet3d} & PointNet & 0.0410 & 0 & 0.8183 & 0.1782 & 8.5261 & 8.0230 \\
HPLFlowNet\cite{hplflownet} & BCL & 0.0041 & 0.0002 & 0.4458 & 0.0960 & 4.3206 & 2.4881 \\
PointRCNN\cite{pointrcnn} & PointNet & \textbf{0.0204} & 0 & 0.5514 & 0.1627 & 3.9888 & 1.6252 \\
LSTM-EM\cite{LSTM-ED} & LSTM & 0.0358 & 0 & 0.3551 & 0.1044 & 1.5885 & 1.0003 \\
\midrule
\midrule
Pillar.M(L\&I)\cite{pillarmotion} & Pillar.E & 0.0245 & 0 & 0.2286 & 0.0930 & 0.7784 & 0.4685 \\
\midrule
MotionNet\cite{motionnet} & STPN & 0.0239 & 0 & 0.2467 & 0.0961 & 1.0109 & 0.6994 \\
\color{gray}MotionNet\cite{motionnet}\textdagger & \color{gray}STPN & \color{gray}\textbf{0.0201} & \color{gray}0 & \color{gray}0.2292 & \color{gray}0.0952 & \color{gray}0.9454 & \color{gray}0.6180  \\
\color{gray}MotionNet\cite{semi-supervised}\textdaggerdbl & \color{gray}STPN & \color{gray}0.0271 & \color{gray}0 & \color{gray}\textbf{0.2267} & \color{gray}\textbf{0.0945} & \color{gray}\textbf{0.8427} & \color{gray}0.5173 \\
\cellcolor{waymolgray}STPN /w(Ours) & \cellcolor{waymolgray}STPN & \cellcolor{waymolgray}0.0228(\textcolor{waymoblue}{\textbf{$\downarrow$4.62\%}}) & \cellcolor{waymolgray}0 & \cellcolor{waymolgray}0.2294(\textcolor{waymoblue}{\textbf{$\downarrow$7.01\%}}) & \cellcolor{waymolgray}0.0946 & \cellcolor{waymolgray}0.9373(\textcolor{waymoblue}{\textbf{$\downarrow$7.28\%}}) & \cellcolor{waymolgray}0.6242\\
\midrule
ST-Trm\cite{ST-Transformer} & STT & 0.0214 & 0 & 0.2426 & 0.0957 & 1.0504 & 0.7247\\
\cellcolor{waymolgray}STT /w(Ours) & \cellcolor{waymolgray}STT & \cellcolor{waymolgray}\textbf{0.0207}(\textcolor{waymoblue}{\textbf{$\downarrow$3.27\%}}) & \cellcolor{waymolgray}\textbf{0} & \cellcolor{waymolgray}\textbf{0.2311}(\textcolor{waymoblue}{\textbf{$\downarrow$4.74\%}}) & \cellcolor{waymolgray}0.0964 & \cellcolor{waymolgray}\textbf{0.9887}(\textcolor{waymoblue}{\textbf{$\downarrow$5.87\%}}) & \cellcolor{waymolgray}\textbf{0.6744} \\
\midrule
STI\cite{BE-STI} & STI & 0.0244 & 0 & 0.2375 & 0.0950 & 0.9078 & 0.6262 \\
\color{gray}BE-STI\cite{BE-STI}\textdagger & \color{gray}STI & \color{gray}\textbf{0.0220} & \color{gray}0 & \color{gray}\textbf{0.2115} & \color{gray}\textbf{0.0929} & \color{gray}\textbf{0.7511} & \color{gray}\textbf{0.5413}  \\
\cellcolor{waymolgray}STI /w(Ours) & \cellcolor{waymolgray}STI & \cellcolor{waymolgray}0.0235(\textcolor{waymoblue}{\textbf{$\downarrow$3.69\%}}) & \cellcolor{waymolgray}\textbf{0} & \cellcolor{waymolgray}0.2268(\textcolor{waymoblue}{\textbf{$\downarrow$4.51\%}}) & \cellcolor{waymolgray}\textbf{0.0928} & \cellcolor{waymolgray}0.8614(\textcolor{waymoblue}{\textbf{$\downarrow$5.11\%}}) & \cellcolor{waymolgray}\textbf{0.5329} \\
\bottomrule
\end{tabular}
\begin{flushleft}
\small
\textbf{Note: }We report the mean errors for static grids, moving grids with speed $\le$ 5 m/s, and moving grids with speed $>$ 5 m/s. Pillar.M(I\&L)\cite{pillarmotion} is the only method trained using both camera and LiDAR modalities. \textdagger: MGDA \cite{MGDA}. \textdaggerdbl: Data augmentation from \cite{semi-supervised}. STPN: MotionNet backbone. STT: ST-Trm backbone. STI: BE-STI backbone.
\end{flushleft}
\vspace{-5pt}

\label{table:basic_experiment}
\end{table*}
\subsection{Instance-enhanced BEV Encoder}
Previous research has struggled with incorporating instance-level knowledge and maintaining instance consistency in motion prediction and classification tasks. To address this limitation, we propose the \textbf{Instance-enhanced BEV Encoder} (IeBE), which integrates instance-level features into the Bird's-Eye View (BEV) representation. This module consists of two stages: early fusion and late fusion. By leveraging detailed object information, it enhances the accuracy and consistency of motion predictions and classifications. We conduct extensive experiments to determine which fusion strategy best suits our task. The final design of our module is illustrated in Fig.~\ref{fig:framework}.

\textbf{Early Fusion.} In order to concentrate significant information in the BEV feature map, a squeeze-and-excitation (SE) layer is incorporated into the BEV features $\mathcal{B}$. The SE attention mechanism is defined as:
\begin{equation}
    \mathcal{F} = \sigma\left(\mathbf{W} f_\text{GAP}(\mathcal{B}\right)) \cdot \mathcal{B}
\end{equation}
Where $\mathbf{W}$ denotes a linear transformation matrix, $f_\text{GAP}$ represents global average pooling, and $\sigma$ is the sigmoid activation function.

Subsequently, the refined BEV feature map $\mathcal{F}$ is flattened into a feature vector, resulting in a feature matrix $\hat{\mathcal{F}} \in \mathbb{R}^{(H\times W) \times D}$. The instance masks embeddings $\mathcal{I} \in \mathbb{R}^{N \times D}$ are first processed through a self-attention layer. To initially incorporate instance-level knowledge into the BEV space. Then, we utilize the Hadamard product and a sigmoid function to update the BEV feature vector $\mathcal{F}'$ using the instance features. This process is formulated as:
\begin{equation}
    \mathcal{F}' = \mathcal{F}
\otimes \sigma\left(\texttt{max}_{t=1}^{N}(\mathcal{F} \otimes \mathcal{I}^{\top})\right)^{\top} 
\end{equation}
After this, BEV features initially have the ability of instance-aware. 

\textbf{Late Fusion.} Then we embed the feature matrix by a learnable linear transformation $\phi$ for BEV feature vectors and instance feature vectors as follows:
\begin{equation}
    X_{i}^* = \phi(X_{i}, {W}_{i}), X_i \in \{{\mathcal{F}'}, \mathcal{I} \}
\end{equation}
Next, we fuse BEV and instance features fully. We perform cross-attention to aggregate instance features $\mathcal{I}^*$ into initially fused BEV features $\mathcal{F}^*$. FFN is used to adjust the feature dimension. 
The late fusion of instance-level features from Interaction-Augmented Instance Encoder is formulated as \ref{eq:fusion}. 

\begin{equation}
\label{eq:fusion}
    \mathcal{E}=\texttt{crossattn}(q=\mathcal{F}^*,k=v=\mathcal{I}^*)
\end{equation}
Where $\mathcal{E}$ represents the final BEV features enhanced with instance-level modeling.
\subsection{Task-specific Heads and Loss Function}

LEGO-Motion is trained using a composite loss function that optimizes cell classification, motion prediction, state estimation and mask prediction with instance-level class. For traditional grid-based task decoder head design, We follow standard practice from MotionNet \cite{motionnet}. 

\textbf{Motion Prediction Loss.} 
To accurately predict the future positions of objects, we employ a weighted smooth L1 loss. This loss ensures that the displacement of each non-empty grid cell is correctly estimated. The motion prediction loss is defined as:
\begin{equation}
    L_{\text{mot}} = \frac{1}{O} \sum_{i=1}^{O} w_i \cdot \text{SmoothL1}(\hat{x}_{\text{mot},i}, x_{\text{mot},i}) 
\end{equation}
where \( \hat{x}_{\text{mot},i} \) represents the predicted displacement for the \( i \)-th cell, \( x_{\text{mot},i} \) is the corresponding ground truth, \( N \) is the total number of non-empty cells, and \( w_i \) balances the representation of different categories by assigning a weight to the \( i \)-th cell.

\textbf{State Estimation Loss.} 
To distinguish between dynamic and static elements in the scene, we use a Cross-Entropy (CE) loss for state estimation. This loss predicts whether each cell is in motion or stationary:
\begin{equation}
    L_{\text{state}} = \frac{1}{O} \sum_{i=1}^{O} w_i \cdot \text{CE}(\hat{x}_{\text{state},i}, x_{\text{state},i}) 
\end{equation}
where \( \hat{x}_{\text{state},i} \) is the predicted motion state of the \( i \)-th cell, and \( x_{\text{state},i} \) is the ground truth. 

\textbf{Cell Classification Loss.} 
For semantic understanding of each grid cell, a cross-entropy loss is used to classify cells into predefined categories. This classification helps the network interpret the scene at a higher semantic level:
\begin{equation}
   L_{\text{cls}} = \frac{1}{O} \sum_{i=1}^{O} w_i \cdot \text{CE}(\hat{x}_{\text{cls},i}, x_{\text{cls},i})
\end{equation}
where \( \hat{x}_{\text{cls},i} \) is the predicted class of the \( i \)-th cell, and \( x_{\text{cls},i} \) is the ground truth class label.


For mask prediction, each instance is represented by a binary mask $\hat{\mathcal{M}}_{i} \in [0,1]^{H \times W}$, obtained via a dot product between the $i$-th instance mask embedding $\mathcal{I}_i$ and the BEV feature maps $\mathcal{B}$, followed by a sigmoid activation. During training, we decode each $\mathcal{I}_i$ into a semantic probability distribution $\hat{p}_i \in \triangle^{K_c+1}$ using a MLP, where $K_c$ categories plus one background class ($\varnothing$) are considered. The loss function of our framework is adapted from MotionNet \cite{motionnet}, with an added mask loss to supervise instance mask predictions.

\textbf{Mask Loss.} The bipartite matching-based assignment \cite{matching} is used to get a matching $\psi$ between the set of prediction and the set of $N^{gt}$ ground truth instance  $\mathcal{I}^{\mathrm{gt}}=\left\{\left(c_{i}^{\mathrm{gt}}, \mathcal{M}_{i}^{\mathrm{gt}}\right) \mid c_{i}^{\mathrm{gt}} \in\{1, \ldots, K_{c}\}, \mathcal{M}_{i}^{\mathrm{gt}} \in\{0,1\}^{H \times W}\right\}_{i=1}^{N^{\mathrm{gt}}}$, where $c_{i}^{\mathrm{gt}}$ is the class of the i-th ground truth instance. Given a matching $\psi$, the mask classification loss $\mathcal{L}_{\text{mask-cls}}$ is composed of a cross-entropy classification loss and a binary mask loss $\mathcal{L}_{\text{mask}}$ for each prediction instance.
\begin{align}
    \mathcal{L}_{\text{mask-cls}}\left(\mathcal{I}, \mathcal{I}^{\mathrm{gt}}\right) 
    &= \sum_{j=1}^{N} \left[ -\log p_{\psi(j)}\left(c_{j}^{\mathrm{gt}}\right) \right. \nonumber \\
    &\quad + \left. \mathbf{1}_{c_{j}^{\mathrm{gt}} \neq \varnothing} \mathcal{L}_{\text{mask}}\left(\mathcal{M}_{\psi(j)}, \mathcal{M}_{j}^{\mathrm{gt}}\right) \right].
\end{align}

The total loss is a weighted sum of individual terms, balancing their contributions during training:
\begin{equation}
    L = \lambda_{\text{mot}} \cdot L_{\text{mot}} + \lambda_{\text{state}} \cdot L_{\text{state}} + \lambda_{\text{cls}} \cdot L_{\text{cls}} + \lambda_{\text{mask-cls}} \cdot L_{\text{mask-cls}}
\end{equation}
where \( \lambda_{\text{mot}} \), \( \lambda_{\text{state}} \), \( \lambda_{\text{cls}} \), and \( \lambda_{\text{mask-cls}} \) are hyperparameters controlling the importance of each loss term.




\section{Experiment}

We conduct our experiments to answer following questions:(1) \textit{Does our proposed learning-enhanced grids with occupancy instance modeling framework improve the capability of motion prediction performance among the traditional grid-based methods?} (2) \textit{How does the instance mask modeling enhanced BEV representation and improve the overall performance?} (3) \textit{Does the motion prediction methods equipped with instance mask modeling capabilities unlock new capabilities such as improved motion stability?}

\begin{figure}[thpb]
      \centering
      \includegraphics[scale=0.55]{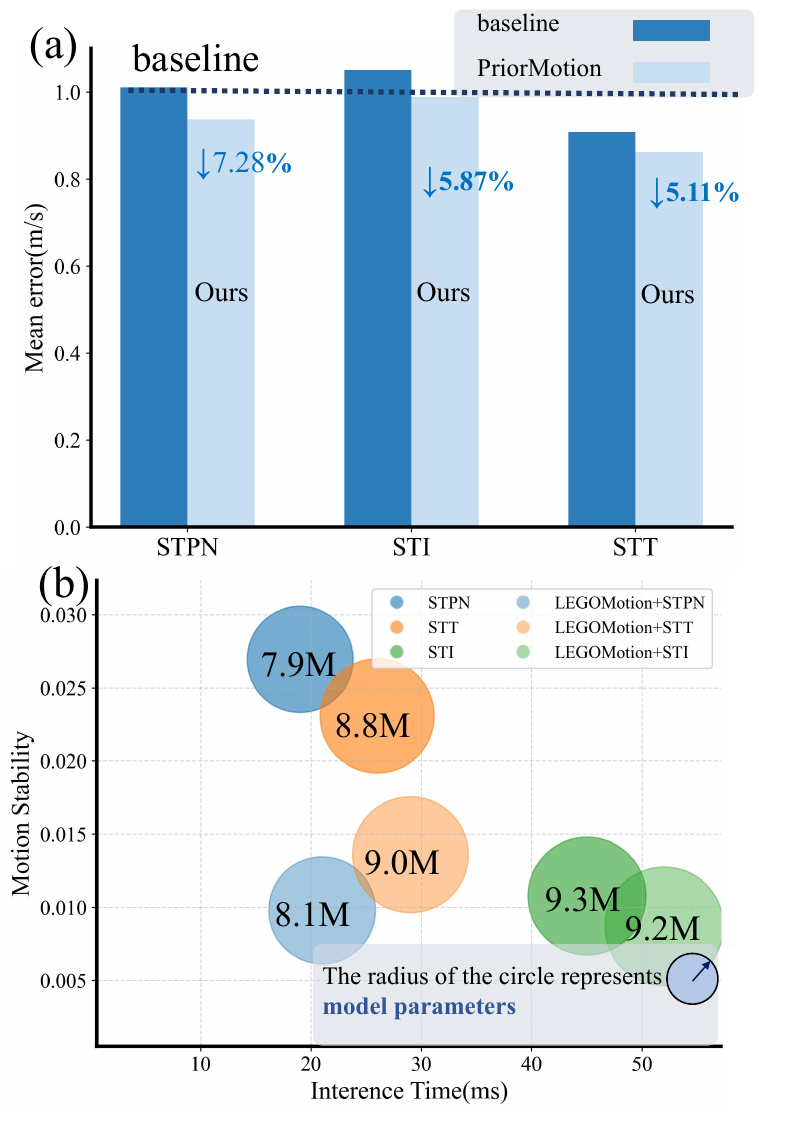}
      \caption{Comparison between grid-based methods with and without LEGO-Motion. The radius of the circle represents the number of model parameters. (a) LEGO-Motion outperforms the baseline in traditional mean speed error; (b) It also demonstrates new capabilities in motion stability, with only a slight increase in model parameters. }
      \label{fig:exp}
   \end{figure}

\begin{figure*}[thpb]
      \centering
      \includegraphics[scale=0.55]{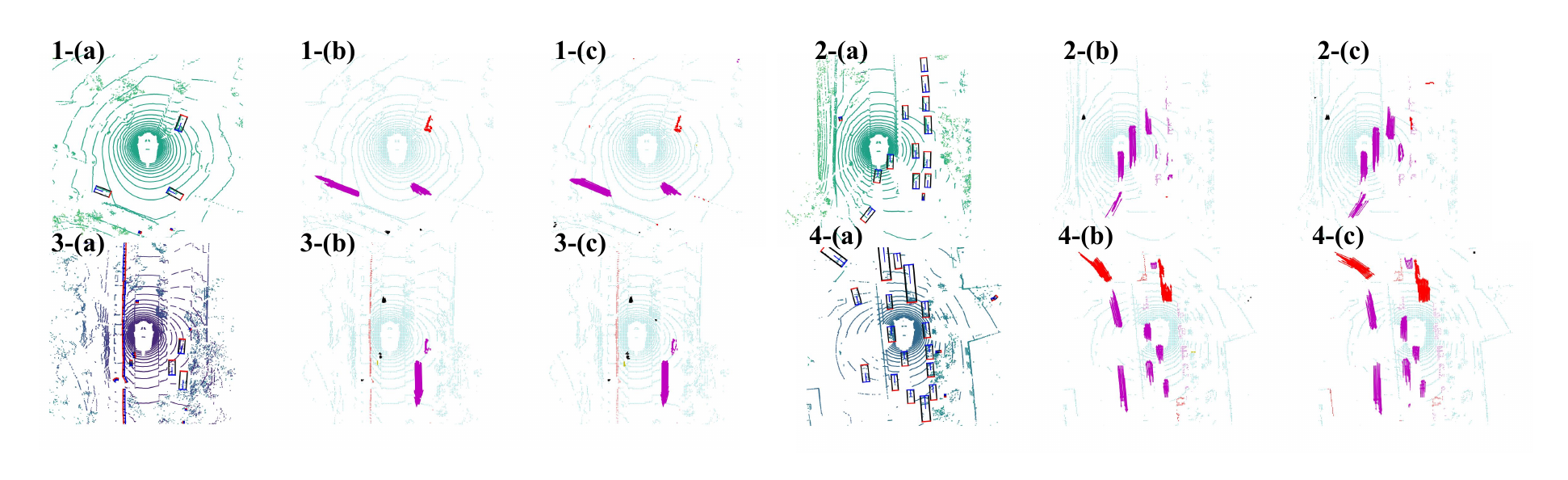}
      \caption{Qualitative results of the proposed LEGO-Motion framework. Each group represents a specific traffic scenario. (a) Object-level ground truth (GT) in Bird's-Eye View (BEV); (b) Grid-level GT; (c) Motion prediction results. Motions are represented by arrows attached to each grid, and cell classification results are indicated by various colors: cyan for background, pink for vehicles, black for pedestrians, yellow for bikes, and red for others.}
      \label{fig:visual}
   \end{figure*}
\subsection{Experimental Setup}
\textbf{Dataset.} We conduct our experiments on the \textbf{nuScenes} dataset \cite{nuScenes}, a large-scale autonomous driving benchmark that provides comprehensive sensor data, including 360-degree LiDAR, cameras, and radars. The dataset comprises 1000 scenes, with 850 scenes allocated for training and validation, and 150 scenes reserved for testing. Following the standard split, we use 500 scenes for training, 100 scenes for validation, and 250 scenes for testing. Each scene lasts approximately 20 seconds, with annotations provided at 2 Hz and LiDAR point clouds captured at 20 Hz.

\begin{table}[htbp]
\centering
\small
\caption{Performance on the auxiliary cell classification task on nuScenes. The backbone used in LEGO-Motion is STPN.}
\resizebox{\linewidth}{!}{
\begin{tabular}{lccccccc}
\toprule
\multirow{2}{*}{Method}& \multicolumn{7}{c}{Classification Accuracy(\%)$\uparrow$}\\
\cmidrule(lr){2-8} & Bg\raisebox{-0.1\height}{\includegraphics[width=0.04\linewidth]{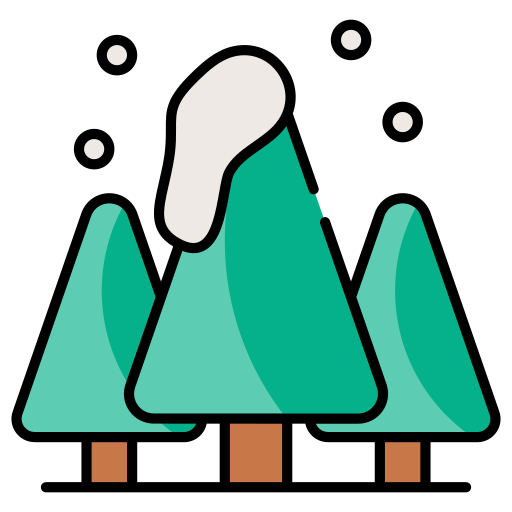}} & Vehicle\raisebox{-0.1\height}{\includegraphics[width=0.05\linewidth]{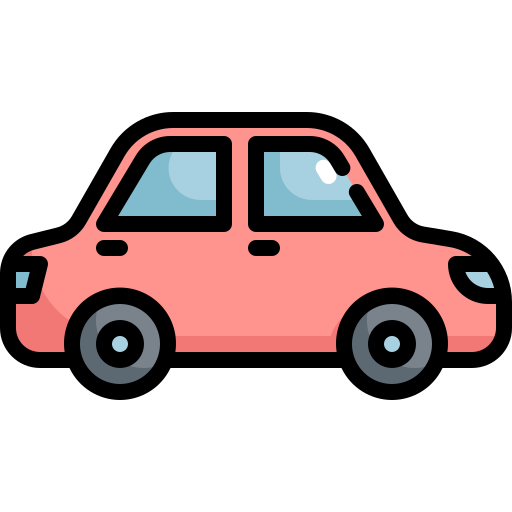}} & Ped.\raisebox{-0.1\height}{\includegraphics[width=0.05\linewidth]{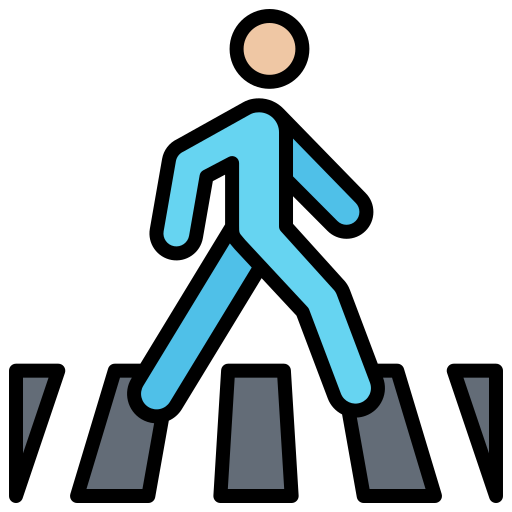}} & Bike\raisebox{-0.1\height}{\includegraphics[width=0.05\linewidth]{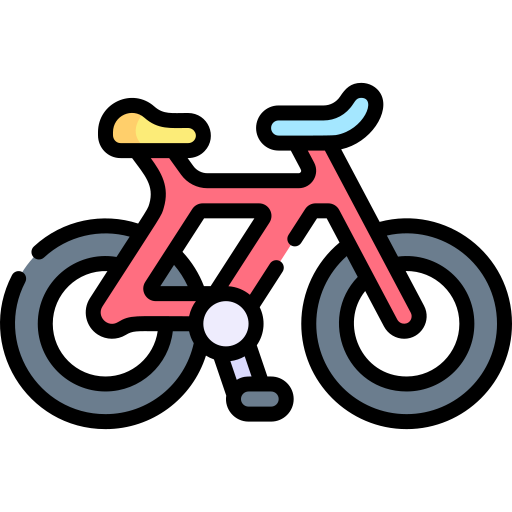}}& Others\raisebox{-0.1\height}{\includegraphics[width=0.05\linewidth]{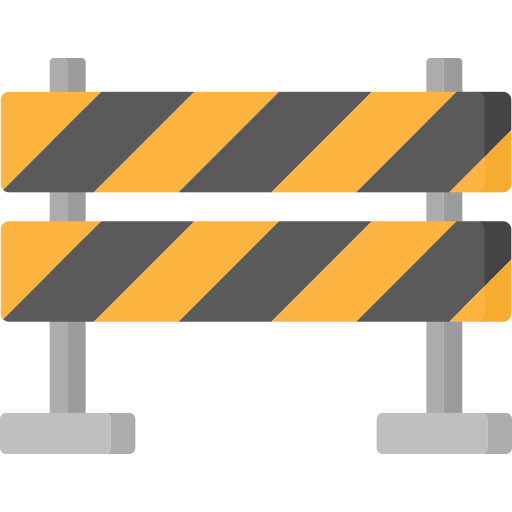}}& MCA & OA \\
\midrule
PointRCNN\cite{pointrcnn}& \textbf{98.4}& 78.7& 44.1& 11.9& 44.0& 55.4 & 96.0 \\
LSTM-ED\cite{LSTM-ED}& 93.8& 91.0& 73.4& 17.9& 71.7& 69.6 & 92.8 \\
\midrule
MotionNet\cite{motionnet}& 97.6& 90.7& 77.2& 25.8& 65.1& 71.3 & \textbf{96.3} \\
\color{gray}MotionNet\textdagger\cite{motionnet} & \color{gray}97.0& \color{gray}90.7& \color{gray}77.7& \color{gray}19.7& \color{gray}66.3& \color{gray}70.3 & \color{gray}95.8 \\
BE-STI\cite{BE-STI}& 97.3& 91.1& 78.6& 24.5& 66.5& 71.6 & 96.0 \\
\color{gray}BE-STI\cite{BE-STI}\textdagger & \color{gray}94.6& \color{gray}\textbf{92.5}& \color{gray}82.9& \color{gray}25.9& \color{gray}77.3& \color{gray}74.7 & \color{gray}93.8 \\
ST-Trm\cite{ST-Transformer}& 96.7& 90.5& 79.0& 21.1& 67.6& 71.0 & 95.5 \\
\cellcolor{waymolgray}LEGO-Motion(Ours)& \cellcolor{waymolgray}96.4& \cellcolor{waymolgray}90.6& \cellcolor{waymolgray}\textbf{81.5} & \cellcolor{waymolgray}\textbf{23.3} & \cellcolor{waymolgray}\textbf{75.2} & \cellcolor{waymolgray}\textbf{73.4} & \cellcolor{waymolgray}95.6 \\
\bottomrule
\end{tabular}
}

\label{table:classification}
\end{table}

\begin{table}
\centering
\small
\caption{Comparison with model parameters, inference time and motion stability results. }
\resizebox{\linewidth}{!}{
\begin{tabular}{lcccc}
\toprule
Method & Backbone & Model P.$\downarrow$ & Infer. T.$\downarrow$ & Motion S.$\downarrow$ \\
\midrule
MotionNet\cite{motionnet} & STPN & 7.9M & 19ms &  0.0267\\
\cellcolor{waymolgray}STPN /w(Ours) & \cellcolor{waymolgray}STPN & \cellcolor{waymolgray}8.1M & \cellcolor{waymolgray}21ms & \cellcolor{waymolgray}\textbf{0.0098} \\
\midrule
STT\cite{ST-Transformer} & STT & 8.8M & 26ms & 0.0231 \\
\cellcolor{waymolgray}STT /w(Ours) & \cellcolor{waymolgray}STT & \cellcolor{waymolgray}9.0M & \cellcolor{waymolgray}29ms & \cellcolor{waymolgray}\textbf{0.0136} \\
\midrule
STI\cite{BE-STI} & STI & 9.2M & 45ms & 0.0108 \\
\cellcolor{waymolgray}STI /w(Ours) & \cellcolor{waymolgray}STI & \cellcolor{waymolgray}9.3M & \cellcolor{waymolgray}52ms & \cellcolor{waymolgray}\textbf{0.0087} \\
\bottomrule
\end{tabular}
}
\label{table:new metric}
\begin{flushleft}
\small
\textbf{Note: }We report the model parameters within our framework integrated with different backbone. \textbf{Model P.}: model parameters. \textbf{Infer. T.}: inference time. \textbf{Motion S.}: motion stability.
\end{flushleft}
\vspace{-10pt}
\end{table}

\textbf{Implementation details.} 
For a fair comparison, we adopt the same data preprocessing pipeline as described in \cite{motionnet}. Input point clouds are cropped to the range of $[-32\text{m}, 32\text{m}] \times [-32\text{m}, 32\text{m}] \times [-3\text{m}, 2\text{m}]$ and voxelized with a resolution of $0.25\text{m} \times 0.25\text{m} \times 0.4\text{m}$. Each sequence consists of 5 frames, where the last frame corresponds to the current time, and the previous 4 frames are from past timestamps. The time interval between consecutive frames is 0.2 seconds.
Our LEGO-Motion framework is implemented in PyTorch and trained in two stages using the Adam optimizer \cite{Adam}. The backbone used is STPN \cite{motionnet} unless otherwise specified. The network is trained for 45 epochs with a batch size of 4 on a single Tesla A100 GPU. We set the initial learning rate to 0.0016, which is decayed by a factor of 0.5 at epochs 10, 20, 30, and 40. During the first 20 epochs, we set $\lambda_{\text{mot}} = \lambda_{\text{state}} = \lambda_{\text{cls}} = 1.0$ and $\lambda_{\text{mask-cls}} = 2.0$. For the final 25 epochs, these parameters are adjusted to $\lambda_{\text{mot}} = \lambda_{\text{state}} = \lambda_{\text{cls}} = 2.0$ and $\lambda_{\text{mask-cls}} = 1.0$. This strategy aims to initially enhance the instance feature extraction capability of LEGO-Motion with spatial acquisition and then shift the focus to grid-based tasks.
\textbf{Evaluation metrics.}
Following previous works \cite{BE-STI}, we categorize non-empty cells into three groups based on their speeds: static ($\mathbf{speed} \leq 0.2 \, \text{m/s}$), slow ($0.2 \, \text{m/s} < \mathbf{speed} \leq 5 \, \text{m/s}$), and fast ($\mathbf{speed} > 5 \, \text{m/s}$).

For each group, we report the mean and median prediction errors, which are calculated as the L2 distance between the predicted displacements and the ground truth displacements 1 second into the future. Additionally, we evaluate performance on auxiliary cell and instance classification tasks. The overall accuracy (OA) is reported as the average accuracy over all non-empty cells:
\begin{equation}
    \text{OA} = \frac{1}{N} \sum_{i=1}^{N} \mathbb{I}(\hat{y}_i = y_i)
\end{equation}

where \( N \) is the total number of non-empty cells, \( \hat{y}_i \) is the predicted class, and \( y_i \) is the ground truth class for cell \( i \).

We also report the mean category accuracy (MCA), which is the average accuracy over all five categories.

\begin{equation}
\text{MCA} = \frac{1}{C} \sum_{c=1}^{C} \frac{\text{TP}_c}{\text{TP}_c + \text{FN}_c}
\end{equation}
where \( C \) is the number of categories, \( \text{TP}_c \) is the number of true positives for category \( c \), and \( \text{FN}_c \) is the number of false negatives for category \( c \).

\begin{table*}
\centering
\small
\caption{Performance comparison of our models with different modular design on nuScenes. Each group listed here are implemented without MGDA\cite{MGDA} for purely evaluation of components.}
\vspace{-5pt}
\begin{tabular}{lccc cc cc cc c}
\toprule
\multirow{2}{*}{Method} & \multicolumn{3}{c}{Module} & \multicolumn{2}{c}{Static} & \multicolumn{2}{c}{Speed $\leq$ 5m/s} & \multicolumn{2}{c}{Speed $>$ 5m/s} & \multirow{2}{*}{Speed Stability} \\
\cmidrule(lr){2-4} \cmidrule(lr){5-6} \cmidrule(lr){7-8} \cmidrule(lr){9-10} 
 & IaIE & Early F. & Late F. & Mean$\downarrow$ & Median$\downarrow$ & Mean$\downarrow$ & Median$\downarrow$ & Mean$\downarrow$ & Median$\downarrow$ & Variance$\downarrow$ \\
\midrule
MotionNet & \ding{55} & \ding{55} & \ding{55} & 0.0239 & 0 & 0.2467 & 0.0961 & 1.0109 & 0.6994 & 0.0267 \\
(a) & \cellcolor{waymollblue}$\checkmark$ & \ding{55} & \ding{55} & 0.0242 & 0 & 0.2417 & 0.0960 & 0.9881 & 0.7032 & 0.0192 \\
(b) & \ding{55} & \cellcolor{waymollblue}$\checkmark$ & \ding{55} & 0.0251 & 0 & 0.2647 & 0.0977 & 1.0542 & 0.7674 & 0.0294 \\
(c) & \cellcolor{waymollblue}$\checkmark$ & \cellcolor{waymollblue}$\checkmark$ & \ding{55} & \textbf{0.0220} & 0 & 0.2391 & 0.0949 & 0.9611 & 0.6502 & 0.0107 \\
(d) & \cellcolor{waymollblue}$\checkmark$ & \cellcolor{waymollblue}$\checkmark$ & \cellcolor{waymollblue}$\checkmark$ & 0.0228 & \textbf{0} & \textbf{0.2294} & \textbf{0.0946} &\textbf{0.9373} & \textbf{0.6242} & \textbf{0.0098} \\
\bottomrule
\end{tabular}
\vspace{-5pt}
\label{table:module_ablation}
\end{table*}

\subsection{Main Results}
\textbf{Comparison with State-of-the-Art methods.}
We compare our proposed LEGO-Motion framework with a variety of published algorithms in Tab. \ref{table:basic_experiment}. For fairly comparison, any training strategies are not used in our reported results. Regarding output format, all published methods can be categorized into object-level and grid-level methods. Our method reports the SOTA result with reference to the mean prediction error on both slow and fast moving objects compared with all object-level methods. By comparing the results of grid-based methods with and without LEGO-Motion, we observe a significant performance improvement when using instance mask information in contrast to models that do not incorporate LEGO-Motion. 
Specifically, Compared to the baseline MotionNet \cite{motionnet}, our learning-enhanced grids gives about 4.62\% boost to the static group, 7.01\% performance boost to the slow speed group, 7.28\% performance boost to the fast speed group. We also report the cell classification results of our proposed method in Tab. \ref{table:classification}. Our method perform higher accuracy on almost all of the foreground objects and MCA. The experimental results show that our framework is a robust way to scene representation and motion tasks.

\textbf{Performance of motion stability.}
We report motion stability metric proposed in PriorMotion \cite{priormotion} with previous methods(shown in Tab. \ref{table:new metric}). Our method show improvements in motion stability metric by comparing the baselines with or without LEGO-Motion. Specifically, our method outperforms the MotionNet with a margin of 0.0269 motion stability. It indicates that LEGO-Motion unlocks the grid-based methods' new capabilities of maintaining motion stability.

\textbf{Runtime analysis.}
We present a latency analysis and model parameter comparison in Table \ref{table:new metric}. The LEGO-Motion framework achieves performance improvements without a significant increase in the number of parameters, effectively balancing performance with computational efficiency. During inference, our entire model, equipped with the STPN backbone, runs in 21 ms on a single RTX 3090 GPU, meeting the requirements for real-time applications. 

\textbf{Qualitative results.}
We show the qualitative results of our LEGO-Motion structure in Fig. \ref{fig:visual}. The predicted motion is represented by an arrow attached to each grid cell, whose length and direction represent the displacement 1s into future. As we can see, LEGO-Motion yields high-quality motion prediction results on BEV grid cells.

\begin{table}
\centering
\vspace{-5pt}
\small
\caption{Comparison results with different fusion paradigms, including: 
(a) concatenation; 
(b) addition; 
(c) weighted sum.}
\vspace{-5pt}
\resizebox{\linewidth}{!}{
    \begin{tabular}{lc cc cc cc}
    \toprule
    \multirow{2}{*}{Method}  & \multicolumn{2}{c}{Static} & \multicolumn{2}{c}{Speed $\leq$ 5m/s} & \multicolumn{2}{c}{Speed $>$ 5m/s} \\
    \cmidrule(lr){2-3} \cmidrule(lr){4-5} \cmidrule(lr){6-7}
     & Mean$\downarrow$ & Median$\downarrow$ & Mean$\downarrow$ & Median$\downarrow$ & Mean$\downarrow$ & Median$\downarrow$ \\
    \midrule
    Base & 0.0239 & 0 & 0.2467 & 0.0961 & 1.0109 & 0.6994\\
    
    
    (a) & 0.0264 & 0 & 0.2724 & 0.1138 & 1.1634 & 0.7516 \\
    (b) & 0.0249 & 0 & \textbf{0.2341} & \textbf{0.0932} & 1.0059 & \textbf{0.6350} \\
    (c) & \textbf{0.0233} & 0 & 0.2412 & 0.1052 & \textbf{0.9916} & 0.6949 \\
    \bottomrule
    \end{tabular}
}
\label{table:fusion_ablation}
\end{table}
\subsection{Ablation Study}

\textbf{Modular design.}
We evaluate the contributions of each component in LEGO-Motion, as shown in Table \ref{table:module_ablation}. An ablation study was conducted to analyze the effectiveness of the IaIE. Comparing configuration (a) with (b), introducing IaIE significantly reduces prediction errors and enhances motion stability. This highlights the importance of interaction relationships for extracting better motion behaviors.

Additionally, we analyzed the designs of our proposed Instance-enhanced BEV Encoder (IeBE), which includes early fusion (EF) and late fusion (LF). EF alone does not fully inject instance features into BEV grid cells without cross-attention, while LF directly uses cross-attention without initial wrappers. Both modules improve motion prediction performance, but their combination further enhances performance significantly. This confirms the importance of the fusion design (Eq.\ref{eq:fusion}) in unlocking the full potential of latent instance-to-BEV enhanced modeling.

\textbf{Fusion design.}
To thoroughly validate our IeBE, we conducted experiments using instance mask prediction results $\mathcal{I} \in \mathbb{R}^{N \times W \times C}$ fused with BEV feature maps. We compared various fusion methods for integrating instance mask features with BEV feature maps, as shown in Table~\ref{table:fusion_ablation}. The weighted sum operation demonstrated similar performance to direct addition, while both methods outperformed the concatenation operation. Besides, it indicates that simple fusion methods may not fully incorporate instance-level features into BEV feature maps.
\begin{table}
\centering
\vspace{-5pt}
\small
\caption{Motion Prediction mean error on FMCW LiDAR benchmark.}
\vspace{-5pt}
\resizebox{\linewidth}{!}{
    \begin{tabular}{lcc cc cc cc}
    \toprule
    \multirow{2}{*}{Method} & \multicolumn{2}{c}{Static} & \multicolumn{2}{c}{Speed $\leq$ 5m/s} & \multicolumn{2}{c}{Speed $>$ 5m/s} \\
    \cmidrule(lr){2-3}\cmidrule(lr){4-5} \cmidrule(lr){6-7}
     & Mean$\downarrow$ & Median$\downarrow$ & Mean$\downarrow$ & Median$\downarrow$ & Mean$\downarrow$ & Median$\downarrow$ \\
    \midrule
    MotionNet\cite{motionnet} & 0.0402 & 0 & 0.5666 & 0.2826 & 1.6428 & 0.5368 \\
    STI\cite{BE-STI} & 0.0538 & 0 & \textbf{0.5153} & 0.3880 & \textbf{1.3689} & 0.5799\\
    STT\cite{ST-Transformer} & 0.0398 & 0 & 0.5580 & \textbf{0.2932} & 1.5798 & 0.5249\\
    STPN /w (Ours) &\textbf{0.0384}& \textbf{0} & 0.5327 & 0.3241 & 1.4653 & \textbf{0.5211} & \\
    \bottomrule
    \end{tabular}
}
\vspace{-5pt}
\label{table:fmcw}
\end{table}

\textbf{Performance on FMCW LiDAR benchmark.}
For further comparison, we implement both MotionNet and LEGO-Motion on our proprietary benchmark. The data is collected using FMCW LiDAR, the most advanced LiDAR technology currently available for autonomous driving vehicles. The dataset comprises 15,000 samples for training, 1,500 samples for validation, and 3,500 samples for testing. Our results show that LEGO-Motion outperforms MotionNet with mean errors of 0.0018 m, 0.0339 m and 0.1775 m for the static, slow-moving and fast-moving object groups, respectively (as shown in Tab. \ref{table:fmcw}). These findings confirm the robustness of LEGO-Motion and highlight its potential for widespread application in real-world scenarios.









\section{Conclusion}
We propose LEGO-Motion, a novel occupancy-instance fusion framework that integrates instance-level semantic features into Bird's Eye View (BEV) grids, enabling unified modeling of both geometric occupancy and interactive motion dynamics. Interaction-augmented Instance Encoder(IaIE) explicitly captures traffic participants' interdependencies through attention mechanisms, and Instance-enhanced BEV Encoder(IeBE) dynamically injects instance semantics into grid-based geometric features. Extensive experiments on the nuScenes dataset demonstrate that LEGO-Motion achieves state-of-the-art performance, surpassing existing occupancy-based methods by 12.7\% in motion accuracy while maintaining real-time inference efficiency. The framework's generalization capability is further validated on the FMCW LiDAR benchmark. These results confirm that our hybrid paradigm successfully combines the strengths of instance-aware reasoning and occupancy-based representation. We anticipate this work will inspire robust instance-grid fusion paradigms for autonomous driving in open-world scenarios

\bibliographystyle{IEEEtranBST/IEEEtran}
\bibliography{ref}

\end{document}